\documentclass[conference]{IEEEtran}
\IEEEoverridecommandlockouts
\usepackage{subfigure}

\usepackage{lipsum}
\newcommand{\RNum}[1]{\uppercase\expandafter{\romannumeral #1\relax}}
\usepackage[a4paper, total={6in, 8in}]{geometry}
\renewcommand{\baselinestretch}{0.9} 

\usepackage{cite}
\usepackage{amsmath,amssymb,amsfonts}
\usepackage{algorithmic}
\usepackage{graphicx}
\usepackage{textcomp}
\usepackage{xcolor}
\def\BibTeX{{\rm B\kern-.05em{\sc i\kern-.025em b}\kern-.08em
    T\kern-.1667em\lower.7ex\hbox{E}\kern-.125emX}}
\begin{document}

\title{Vision-Based Layout Detection from Scientific Literature using Recurrent Convolutional Neural Networks}

\author{\IEEEauthorblockN{Huichen Yang, William H. Hsu}
\IEEEauthorblockA{\textit{Department of Computer Science} \\
\textit{Kansas State University}\\
Manhattan, Kansas, USA \\
huichen@ksu.edu, bhsu@ksu.edu}
}

\maketitle

\begin{abstract}
We present an approach for adapting convolutional neural networks for object recognition and classification to scientific literature layout detection (SLLD), a shared subtask of several information extraction problems. Scientific publications contain multiple types of information sought by researchers in various disciplines, organized into an abstract, bibliography, and sections documenting related work, experimental methods, and results; however, there is no effective way to extract this information due to their diverse layout. In this paper, we present a novel approach to developing an end-to-end learning framework to segment and classify major regions of a scientific document. We consider scientific document layout analysis as an object detection task over digital images, without any additional text features that need to be added into the network during the training process. Our technical objective is to implement transfer learning via fine-tuning of pre-trained networks and thereby demonstrate that this deep learning architecture is suitable for tasks that lack very large document corpora for training \textbf{\textit {ab initio}}. As part of the experimental test bed for empirical evaluation of this approach, we created a merged multi-corpus data set for scientific publication layout detection tasks. Our results show good improvement with fine-tuning of a pre-trained base network using this merged data set, compared to the baseline convolutional neural network architecture. 

\end{abstract}

\begin{IEEEkeywords}
Object Detection, Deep Learning, Transfer Learning, Information Extraction, Document Layout Analysis  
\end{IEEEkeywords}

\section{Introduction}
The number of academic literature publications has been growing rapidly. These published literature documents include a great amount of free text containing potentially valuable information which can help scientists and researchers to develop new ideas in their fields of interest, to extract peers' key insights \cite{b1}, or explore new research areas from various fields each having their own serials (journals) and conference proceedings \cite{b2}. Unfortunately, the increasing rate of production of scientific literature in many domains has outpaced the rate at which individual researchers and smaller laboratories can process new documents and assimilate knowledge. Millions of academic products (conference papers, journal papers, book chapters, etc.) are published each year and disseminated as digital documents consisting of generally unstructured text. Scientific Literature Layout Detection (SLLD) can increase the effectiveness of automated information extraction tools by inferring the layout of scientific literature accurately based on geometric and logistic layout analysis, this facilitating the automatic extraction of metadata such as section delimiters, formatting information for equations and formulas, procedural data, special environments, captions for tables and figures, bibliographies, etc. It presents a possible solution for automatic construction of large corpus, and also provides assistance for some scientific literature-related downstream tasks of natural language processing (NLP).

\begin{figure}[h!]
    \begin{center}
        \includegraphics[width=0.3\textwidth]{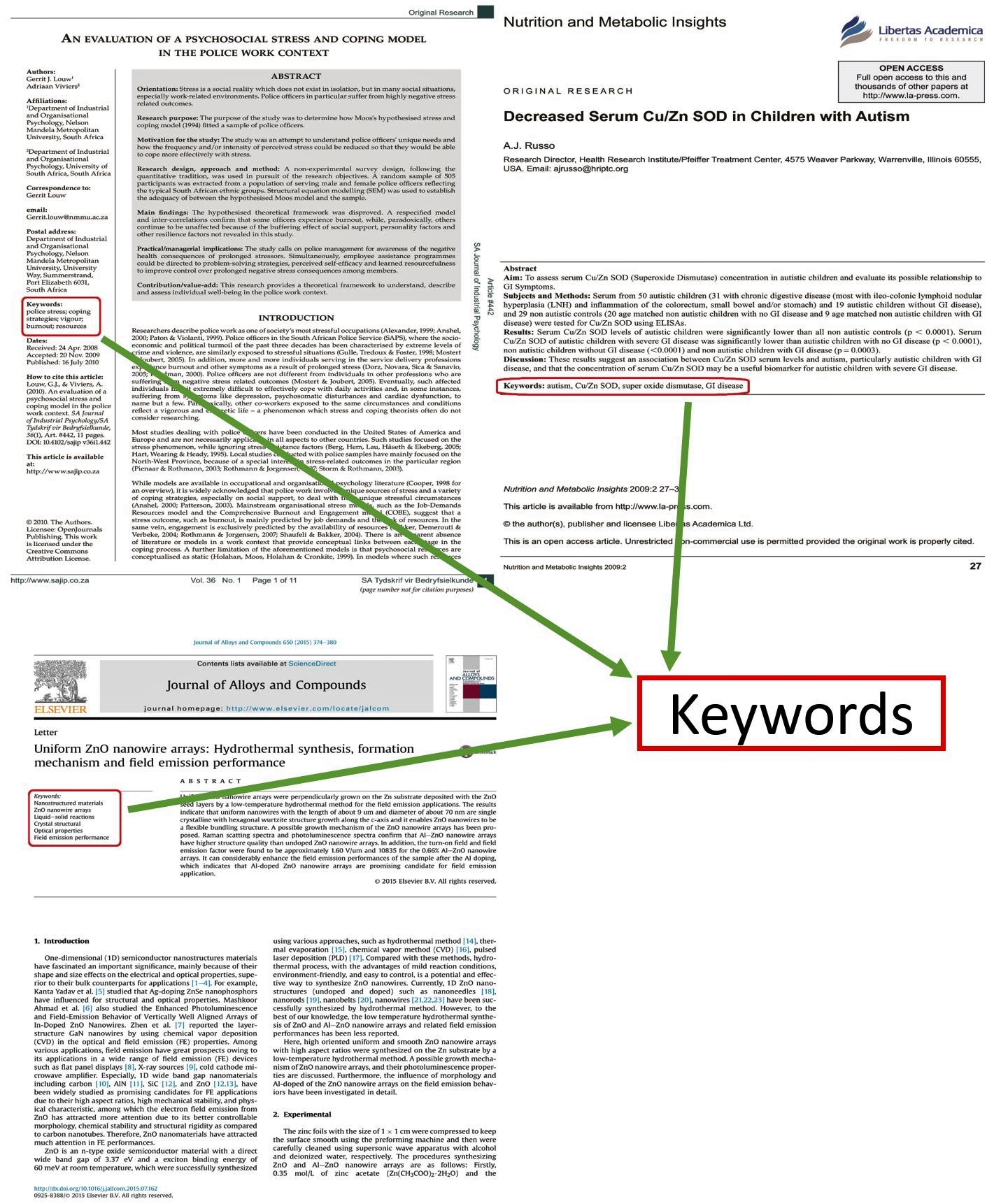}
    \end{center}
\caption{Examples of inconsistent layouts in scientific literature. The layouts of keyword vary by different articles}
\end{figure}

Portable Document Format (PDF) is a form of digital document which has been most commonly used in scientific publications. Extracting good quality metadata from PDF publications remains difficult and challenging, since publishers have very diverse preferences of formatting and layouts in their articles. Thus, a single type of information can be organized in various different formatting styles and fonts for different scientific publications (Fig. 1). Existing tools such as optical character recognition (OCR) perform the extraction of raw text from scanned PDF documents automatically, but they are not used for metadata information extraction, and the lack of layout analysis will lead to messy results, such as the extraction of headers and footers together. Furthermore, raw text information extraction from whole documents tends to necessitate secondary data cleaning, which is extremely time-consuming. For instance, procedural information extraction in the nanomaterial synthesis domain may only require textual information from the experimental methods section rather than an entire scientific article \cite{b1}. In addition, the bottleneck of some existing works of document layout analysis is related to a lack of comprehensive extraction for different blocks, such as the title, authors, figures, etc., from scientific literature, including figure and table detection from digital documents \cite{b4}, full-text extraction from scientific publications \cite{b5}, or domain-specific figure analysis in scientific articles \cite{b6}. Therefore, a robust method for comprehensive and efficient information extraction is still needed for common section types that are typical for many scientific disciplines.

In this paper, we present an end-to-end learning framework that is based on Faster R-CNN \cite{b7}, adapting the two-stage object detection framework from computer vision for scientific literature documents layout detection. This novel approach detects the main regions of scientific articles, and outputs the blocks and their corresponding labels, including title, authors, abstract, body of text, etc.(Fig. 2). We also create a synthesis data set by merging and rendering key pages from two scientific document corpora.  This allows training and evaluation of models that are challenging to evaluate given the insufficiency of existing data set for SLLD tasks. We begin by reviewing related work in Section II and introducing the methodology in detail in Section III. We then fully describe an experiment design in Section IV, present evaluation part in Section V, and finally draw conclusions and priorities for future work in Section VI.

\begin{figure}[h!]
    \begin{center}
        \includegraphics[width=0.5\textwidth]{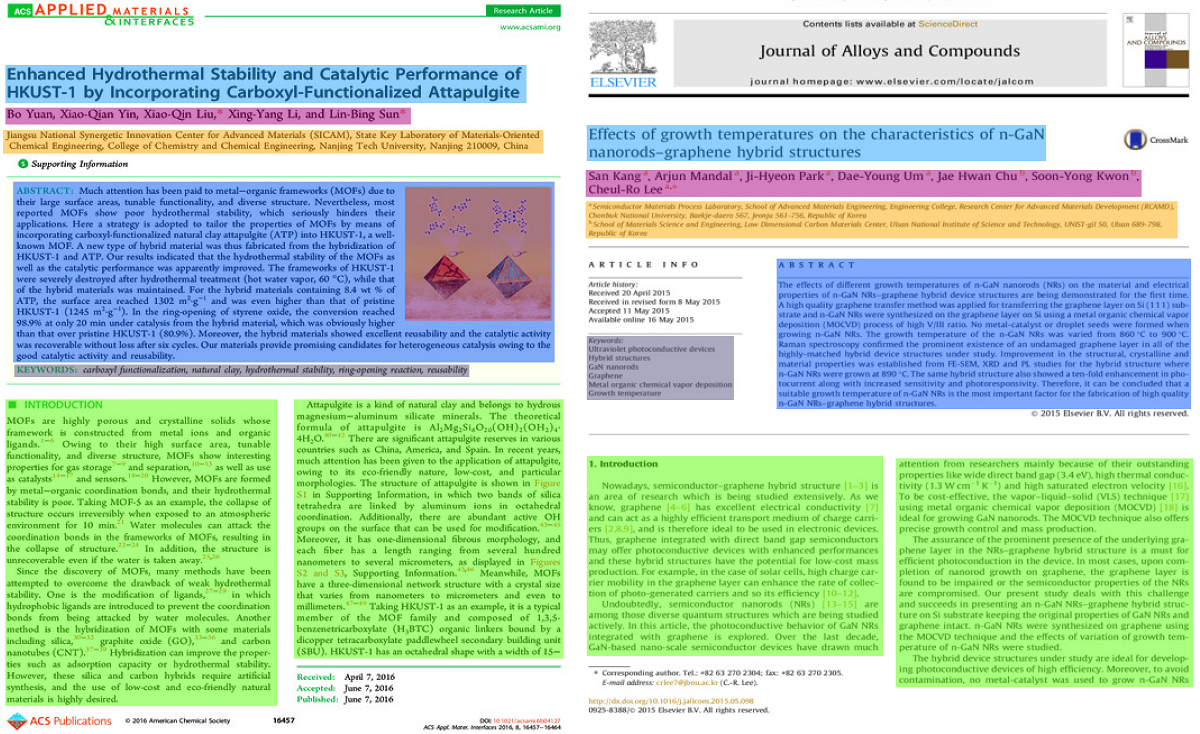}
    \end{center}
\caption{Ground truth: major regions have been annotated}
\end{figure}

\section{Related Work}

For different types of document layout analysis, there are two common ways that have been proposed and contributed by many researchers: rule-based and machine learning-based.

\subsection{Rule-based methods}
Rule-based methods for document layout analysis commonly use regular expressions and heuristic algorithms. They can be divided into three categories: top-down approaches, bottom-up approaches, and hybrid approaches \cite{b8}. The top-down (model-driven) methodology is based on the document layout whose structure is known in advance. Each page of the document will be processed and segmented from large components into smaller sub-components recursively. For instance, a page of scientific literature document with two columns will be split into two blocks of text; next, each block is split into several sectional blocks; then, each sectional block is split into text lines, and so on. Bottom-up (data-driven) methodology is based on of document layout whose structure is unknown in advance. Beginning with clustering document image pixels into relevant components, this method can merge characters into words, lines, and zones, such as CERMINE, an open source tool that extracts metadata information from academic publications \cite{b21}. Hybrid is the method that mixes top-down and bottom-up approaches. Kruatrachue et al. described a hybrid approach which uses edge following algorithm with small window of 16 by 32 pixels to scan a page of document, then uses X-Y cut algorithm to reduce errors if the space is smaller than the window for document segmentation \cite{b9}.

Rule-based methods played an important rule in document layout analysis before machine learning algorithms became popular. However, rule-based methods are based on knowledge of document structure, good feature selection, and sensitive to noise of input document formatting. Such traditional knowledge-based systems could be limited in generality of purpose and robustness across different use cases if the rules are too domain-specific, or quantitatively brittle and arbitrary.

\subsection{Machine learning-based method}
Machine Learning methodologies are also proposed for document layout analysis. The methodologies are generally divided into the following approaches: non-deep and deep learning \cite{b10}.

\textbf{\textit{Non-deep learning}}
Non-deep learning is usually built on conventional machine learning model which is either pixel-based or feature-based. For example, Marinai et al. uses pixel-based Artificial Neural Network (ANN) for document layout analysis \cite{b11}, Wei et al. introduces feature-based Support Vector machines (SVM) \cite{b12}. Usually, pixel-based methods are not a good choice compared with feature-based methods because they may raise missing context information issues [13]. Feature-based methods require feature extraction to empower training and build robust models. These features can be either handcrafted or generated aromatically, such as the cases of texture features extraction \cite{b14} and geometric features extraction \cite{b15} methods for text-line extraction tasks. 
 
\textbf{\textit{Deep learning}}
As a deep learning method, convolutional neural networks (convnets) have become a preeminent architecture for many pattern recognition and computer vision tasks since convolutional neural networks were first successfully used to recognize handwritten characters and eventually adapted to object detection and recognition (particularly the PASCAL VOC and ISLVRC ImageNet challenges). Three key strengths of convnets include differentiable representation, scalable GPU computing, and large data set availability (a resource that is notably lacking at present in the SLLD domain). Models that are trained using deep learning methods can address more complex layout document for both physical layout and logical layout analysis \cite{b16, b17, b3}. For example, Grüning et al. \cite{b18} presented ARU-Net deep neural network as an extension of the U-net \cite{b19} to fix the pooling issue of the previous deep learning methods for text-line detection in historical documents, and Yang et al. \cite{b20} addressed an end-to-end multimodal to extract semantic structures from document images with text features based on a fully convolutional network. Deep neural networks also need large data sets to learn crucial parameters for segmentation or classification tasks. Those parameters can be initialized with random weights, or using transfer learning with pre-trained networks.

\begin{figure*}[!t]
\begin{center}
\includegraphics[height=0.3\textwidth]{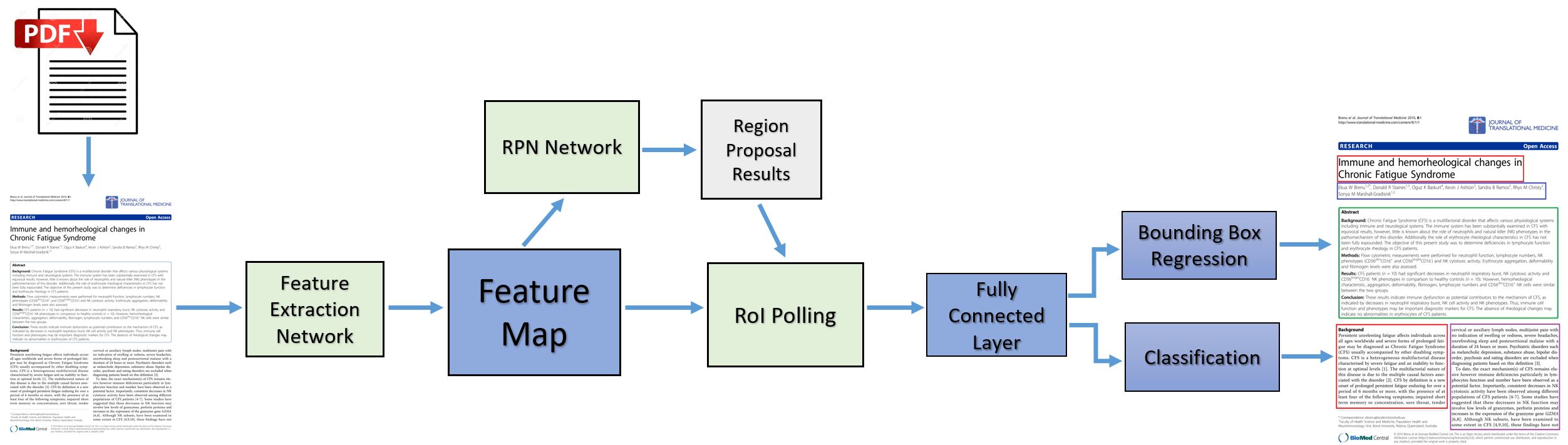}
\end{center}
\caption{Scientific literature layout detection framework}
\end{figure*}

\section{Layout Detection Methodology}
Object detection using deep learning networks has grown popular and for the past decade convnets have held first position and attained state-of-the-art results on standard data sets for object detection (e.g. MS COCO \cite{b22}). The task of SLLD is similar to that of object detection, both of which can locate objects with bounding boxes, and classify these objects in images with labels. However, digital images in scientific publications are not like the normal images that include distinct objects like a car, a bird, or a flower. Most of the scientific document pages only have a few images or tables, and the rest of the parts are the body text of sections, titles, author lists and affiliations, etc., and the types of font tend to be different from each other. To address this problem, we try to use different neural networks for feature extraction from input images, and to combine them with appropriate anchor ratios for SLLD tasks. Section IV shows significantly improved results generated by our approach. Moreover, our end-to-end learning framework is more robust compared with others (e.g., complex document layout analysis \cite{b21}), and it can be applied to any language without additional features (e.g. text embedding \cite{b23}) but document page image for input. 

Our approach uses Faster R-CNN \cite{b7} (Fig. 3) as our baseline, a classic two stage object detection framework that is built on Fast R-CNN \cite{b24}. The main improvement of Faster R-CNN is to replace a selective search algorithm with a region proposal network, which is based on anchor boxes, to generate proposals for a detection network. Therefore, there are two neural networks that are involved in the training process: one neural network is used for proposal region generation which might include target object, and the other is used for selected regions classification and object detection. Faster R-CNN as our baseline framework uses ResNet-50 \cite{b25} with Feature Pyramid Networks (FPN) \cite{b26} as backbone neural networks for feature extraction and potential region selection. These selected regions are then fed to the second neural network for object classification and bounding box regression.

The two strategies are applied based on baseline of Faster R-CNN: backbone network replacement, and appropriate anchor ratio selection for the SLLD task. The experiments demonstrate that the results are indeed improved using our strategies. 

\subsection{Backbone replacement with VoVNet-v2 \cite{b33}}
At present, deep learning-based object detection models rely on Convolutional Neural Network (CNN) as feature extractors, such as ResNet for Faster R-CNN and DarkNet for YOLOV3 \cite{b32}. Although deep neural networks have performed well on feature extraction, it also incurs the problems of high computational cost and slow training speed. VoVNet \cite{b34} network is proposed to resolve the efficiency problem, and it has been known to have better performance than other deep neural networks based on many experimental results. VoVNet is built on One-Shot Aggregation (OSA) modules which aggregate concatenation feature only once in the last feature map rather than aggregate previous feature at every subsequent layers, for instance, by DenseNet \cite{b35}. VoVNet has several architectures based on different numbers of CNN and OSA on different layers, such as VoVNe-39 and VoVNet-57. VoVNet-v2 further improves performance and efficiency of VoVnet by adding (1)  residual connection which enables us to train deeper networks, such as VoVNetV2-99; (2) Squeeze-and-Excitation (eSE) attention module on the last feature layer to improve the performance. We use VoVNetV2-39 as a backbone for feature extraction within Faster R-CNN framework. 

\subsection{Anchors Aspect Ratio Selection}
Anchor boxes use a similar mechanism as sliding-window to capture the most likely regions which contain objects with different scales and aspect ratios on RPN stage. Different objects have different aspect ratios (width/height) to be detected in images, such as the aspect ratio of a car, which is around 1:2, or that a utility pole, which might be 1:10 or lower. To accurately identify the proper schemes of different anchor sizes of pre-detected objects will improve the prediction results. In SLLD tasks, for instances, the aspect ratio box of text body is different from the aspect ratio box of author. We analyze the distribution of bounding box sizes based on the ground truth object bounding box coordinator of our synthesis data set with K-means cluster anchor box selection \cite{b36}, and got the aspect ratios for different blocks ranging from 0.1 to 4.0 (Fig. 4) by choosing 50 clusters. The details of Anchor parameters configuration are presented in Implemented Details.

\begin{figure}
    \centering
        \includegraphics[width=0.45\textwidth]{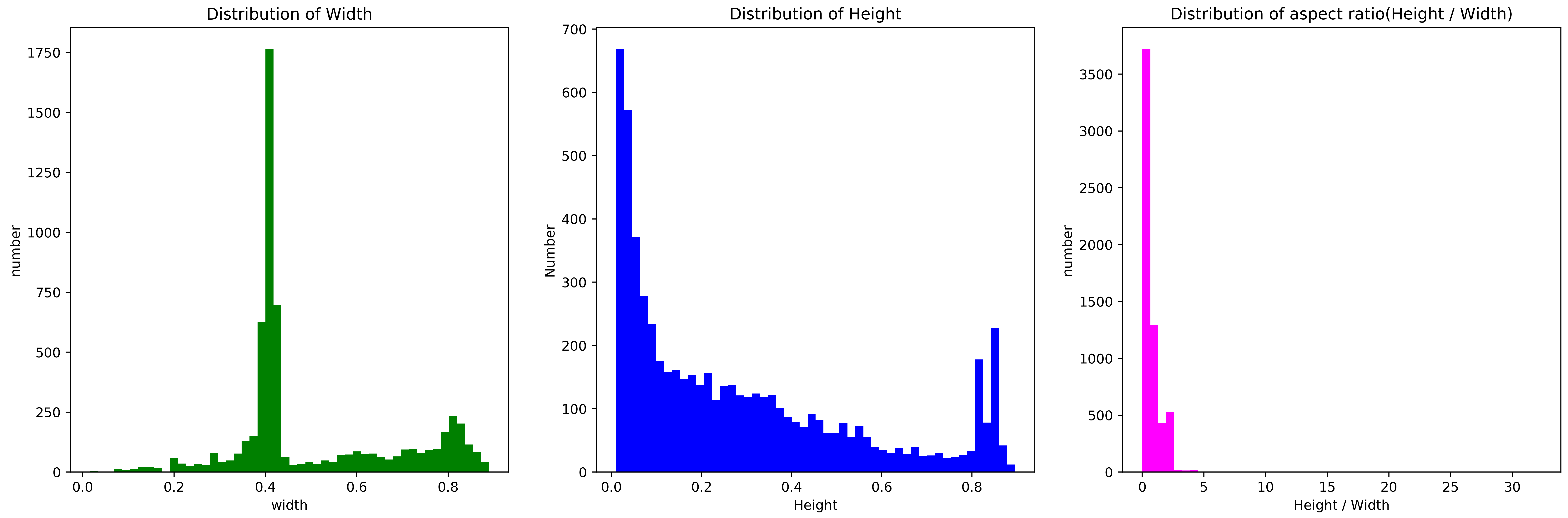}
\caption{Aspect ratio analysis for anchor box}
\end{figure}

\section{Experiment}

\subsection{Synthesis data set}
There are few available data set for SLLD task. Soto et al. introduced a good but small data set that is annotated manually for SLLD \cite{b28}. This data set incldues 822 images from 100 PDF scientific literature, and 9 labeled region classes to cover major region layout of scientific literature. However, it has an instance imbalance issue such that there are 1275 instances of body, which are way more than the 100 instances of title. This is because a paper normally has only one title, a few authors, but the number of body of text are at least as many as the number of pages, and sometime even twice or three times more than the number of pages depending on the layout of the document. Moreover, this data set missed the label of Keywords which usually indicates the most refined and significant information from the scientific literature. In order to solve the above issue, we create a synthesis data set, which is not only an extension of the data set from Soto et al. \cite{b28} but also integrates two other data sets as follows:

\begin{itemize}
    \item ICDAR-2013 \cite{b29}: This data set includes 150 tables from 67 PDF documents, 40 PDFs among which are collected from US Goverment and the rest are from EU. The purpose of constructing such a data set is to increase the diversity of tables in use. All of the PDF documents are converted to images. 
    
    \item GROTOAP \cite{b30}: This data set has 113 annotated PDF documents from scientific literature. It achieves 100\% accuracy since it is annotated manually. We only chose the first page of each PDF scientific document, and converted them into images for our synthesis data set in order to increase the minority instances and solve the instance imbalance issue.
\end{itemize}

\begin{figure}[h!]
    \begin{center}
        \includegraphics[width=0.48\textwidth]{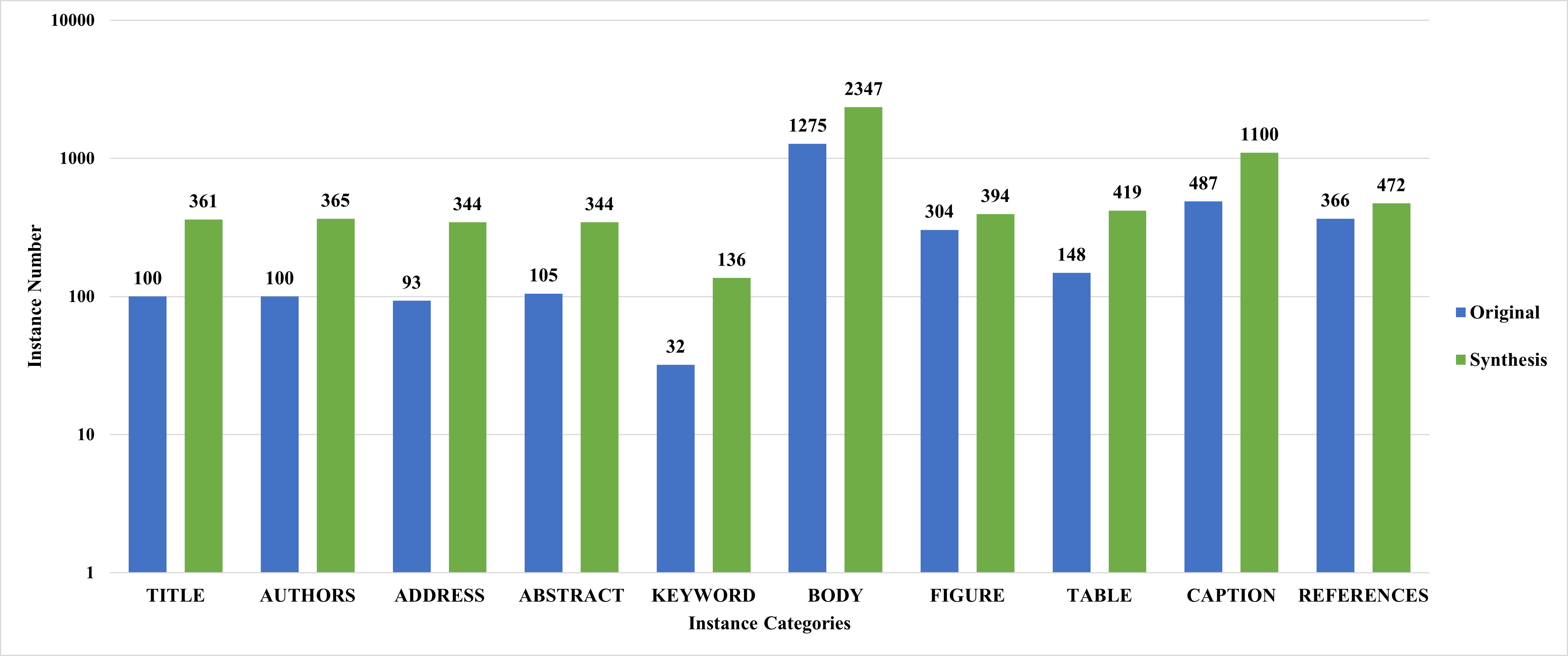}
    \end{center}
\caption{Instances comparison between two data sets by labels}
\end{figure}

After combining these three data sets, we extracted 1550 image pages from 363 PDF documents. All of images have been converted to a fixed size of $612 \times 729$ at 200 dpi. We use 10 labels to classify the major regions from the scientific literature with the synthesis data set:

\begin{itemize}
    \item \textbf{\textit{Title}}: the title and subtitles.
    \item \textbf{\textit{Authors}}: the author names.
    \item \textbf{\textit{Address}}: the affiliation information of authors, including authors' address, email, etc.
    \item \textbf{\textit{Abstract}}: an abstract section.
    \item \textbf{\textit{Keyword}}: the selected keywords.
    \item \textbf{\textit{Body}}: the main block of articles.
    \item \textbf{\textit{Figure}}: all figures but excluding logos or icons from publishers.
    \item \textbf{\textit{Table}}:  the tabular contents.
    \item \textbf{\textit{Caption}}: the captions for both figures and tables
    \item \textbf{\textit{Reference}}: the bibliography information, excluding post-references notes.
\end{itemize}

We added one more label \textit{keyword}, merged \textit{table\_caption} and \textit{figure\_caption} as \textit{\textbf{caption}} from Soto  et  al. [28] data set, and keep the rest as the same. Fig. 5 shows instances comparison between Soto et al. [28] data set and our Synthesis data set. 

\begin{table*}[t]
\caption{Overall comparison among SLLD results (\%) with different methodologies and data set at IoU = 0.5:0.05:0.95. Our detector of Faster RCNN+VoVNetV2-39 performs better than the baseline of Faster RCNN+ResNet50\_FPN. D1 represents data set1 and D2 represents data set2.}
\begin{center}
\begin{tabular}{*{11}{c}}
\hline
 \textbf{Detector} & \textbf{Backbone} & \textbf{Data Set} & \textbf{mAP} & \textbf{AP50} & \textbf{AP75} & \textbf{APs} & \textbf{APm} & \textbf{APl} & \textbf{AR}\\
 \hline
Soto et al.(30 epochs) [28] & ResNet101 & D1 & - & 70.30 & - & - & - & - & - \\
\hline
Faster R-CNN \textbf{(baseline)} & ResNet50\_FPN & D1 & 69.76 & 87.46 & 76.49 & - & 51.65 & 77.41 &  62.70 \\
\hline
Faster R-CNN & ResNet50\_FPN & D2 &77.48& 92.39& 84.42& 35.00 & 63.32 & 77.65  & 69.50 \\
\hline
Mask R-CNN & ResNet50\_FPN & D1 & 70.68 & 87.60 & 82.90 & - & 52.05 & 75.05  & 65.50 \\
\hline
Mask R-CNN & ResNet50\_FPN & D2 & 77.66 & 91.79 & 85.80 & 40.00 & 64.378 & 75.604  & 69.50 \\
\hline
YOLOV3 (49 epochs) [28] & - & D1 & - & 68.90 & - & - & - & -  & - \\
\hline
YOLOV3 & DarkNet53 & D2 &45.90 & 66.50 &57.10 & - & - & -  & 46.33\\
\hline
Faster R-CNN & VoVNetV2-39 & D1 &67.12&89.01 &72.84 & - &47.66 &73.56 &60.50\\
\hline
Faster R-CNN \textbf{(ours)} & VoVNetV2-39 & D2 &76.39 &\textbf{95.02} & \textbf{86.46} & \textbf{75.00} & 62.25 & 74.22 & 68.80 \\
\hline
\end{tabular}
\end{center}
\end{table*}

\begin{figure*}
\begin{center}
  \includegraphics[width=\textwidth,height=8cm]{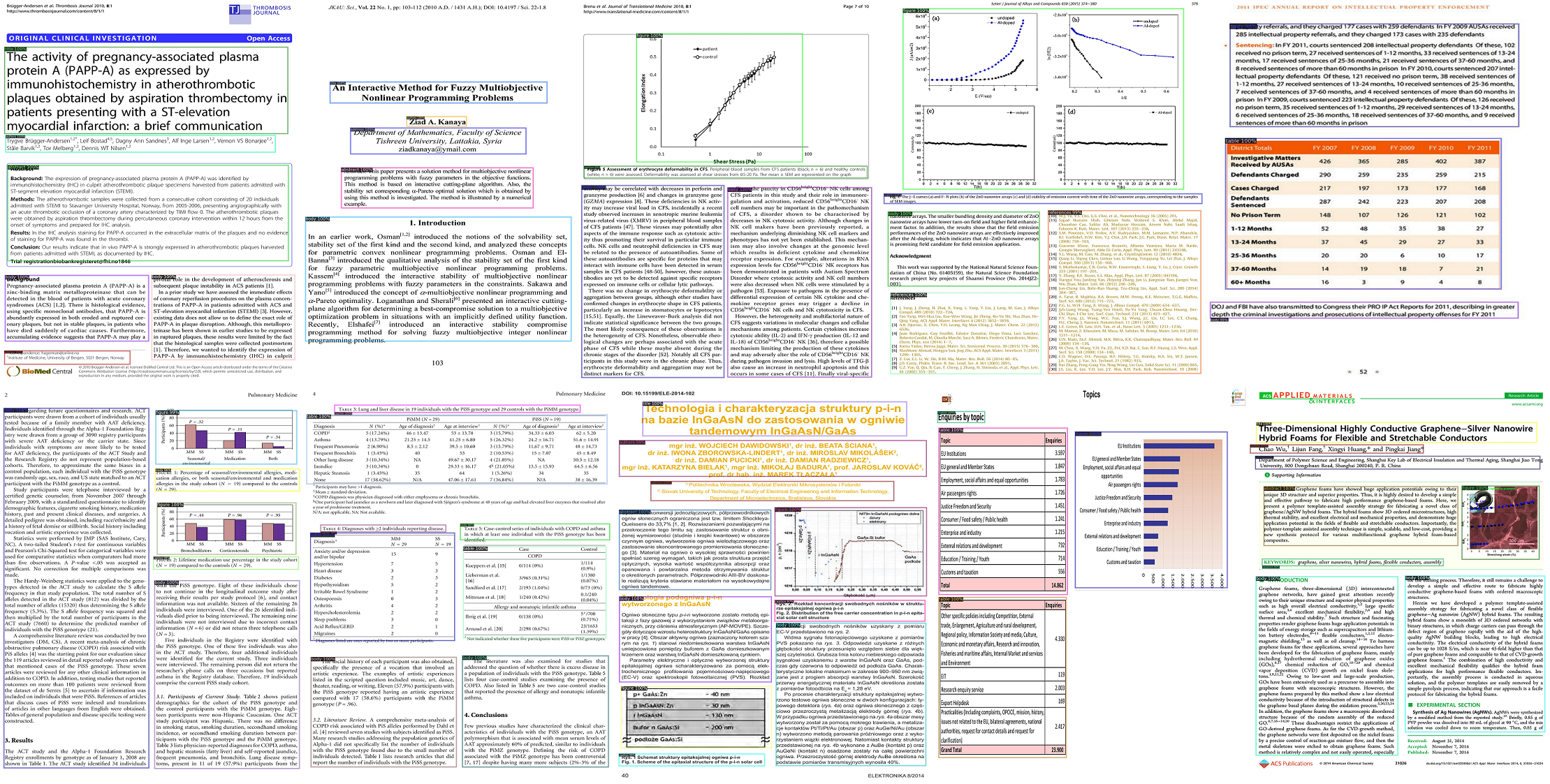}
  \end{center}
  \caption{Detection results with corresponding labels}
\end{figure*}

\subsection{Implementation Details}
The models are trained with pre-trained weights on MS COCO data set ($\sim$37 epochs) \cite{b22}. Two stage object detection frameworks Faster R-CNN and Mask R-CNN are implemented using Detectron2 \cite{b31}. 
All models are trained and tested on a single NVIDIA Tesla P100 GPU with a batch size of 8 for 150 epochs. SGD (Stochastic Gradient Descent) is used as the optimization algorithm. The initial learning rate is 0.002, and decays by 0.1 after 100 epochs. We used different neural networks as a backbone for feature extraction from input images: ResNet-50+FPN, VoVNetV2-39. There are 5 anchor scales in powers of 2 from 32 to 512, 8 anchor aspect ratios from 0.2 to 2.8 based on K-means selection results for VoVNet-v2 backbone model, and the reset of models are trained by standard aspect ratios [0.5, 1.0, 2.0]. We also trained a single-stage object detection framework YOLOv3 \cite{b32} with our synthesis data set for comparison. This model uses DarkNet53 base network that were pre-trained on MS COCO data set.

\section{Evaluation}

We constructed object detection frameworks with different configurations, and train them with two different data set to compare them with our design SLLD model. All of these models start by being trained using pre-trained models on MS COCO \cite{b22} data set. For obtaining a thorough performance evaluation, we use MS COCO evaluation metrics to measure SLLD results rather than other simple evaluation methods, such as precision and recall, because it can help to evaluate various sizes of objects from detection results. For SLLD tasks, it can help to evaluate different region sizes of scientific literature based on the detection results.

\subsection{MS COCO [22] evaluation metrics with different IoU thresholds}
\begin{itemize}
    \item mean Average Precision (\textbf{mAP}): mean average precision at IoU = 0.5:0.05:0.95
    \item Average Precision 50 (\textbf{AP50}): {AP}\textsuperscript{IoU=0.50}
    \item Average Precision 75 (\textbf{AP75}): {AP}\textsuperscript{IoU=0.75}
    \item Average Precision for small object (\textbf{APs}): area (pixel-wise) $<$ 32$^2$ 
    \item Average Precision for medium object (\textbf{APm}): 32$^2$ $<$ area (pixel-wise) $<$ 32$^2$
    \item Average Precision for large object (\textbf{APl}): area (pixel-wise) $<$ 96$^2$
    \item Average Recall (\textbf{AR})
\end{itemize}

\subsection{data set}

\begin{itemize}
    \item \textbf{\textit{data set1}}(D1): original data set \cite{b28} - 600 image for training, 222 images for testing.
    \item \textbf{\textit{data set2}}(D2): synthesis data set - 1225 images for training, 325 images for testing.
\end{itemize}

Our algorithm of SLLD tasks has significantly improved results compared with others (TABLE \RNum{1}), especially for the object detection in small area, such as the area displaying keyword and the area presenting authors in scientific literature. We believe that this result is benefited by our appropriate anchor analysis. We redesign aspect ratio of anchor box rather than using simple numbers [0.5, 1.0, 2.0] for the most commonly used detection tasks to cover more labeled regions in scientific literature documents. Likewise, the result indicates that our framework also obtains improvements in each labeled region of SLLD tasks (Fig. 7).

\begin{figure}[hbt!]
    \begin{center}
        \includegraphics[width=0.48\textwidth]{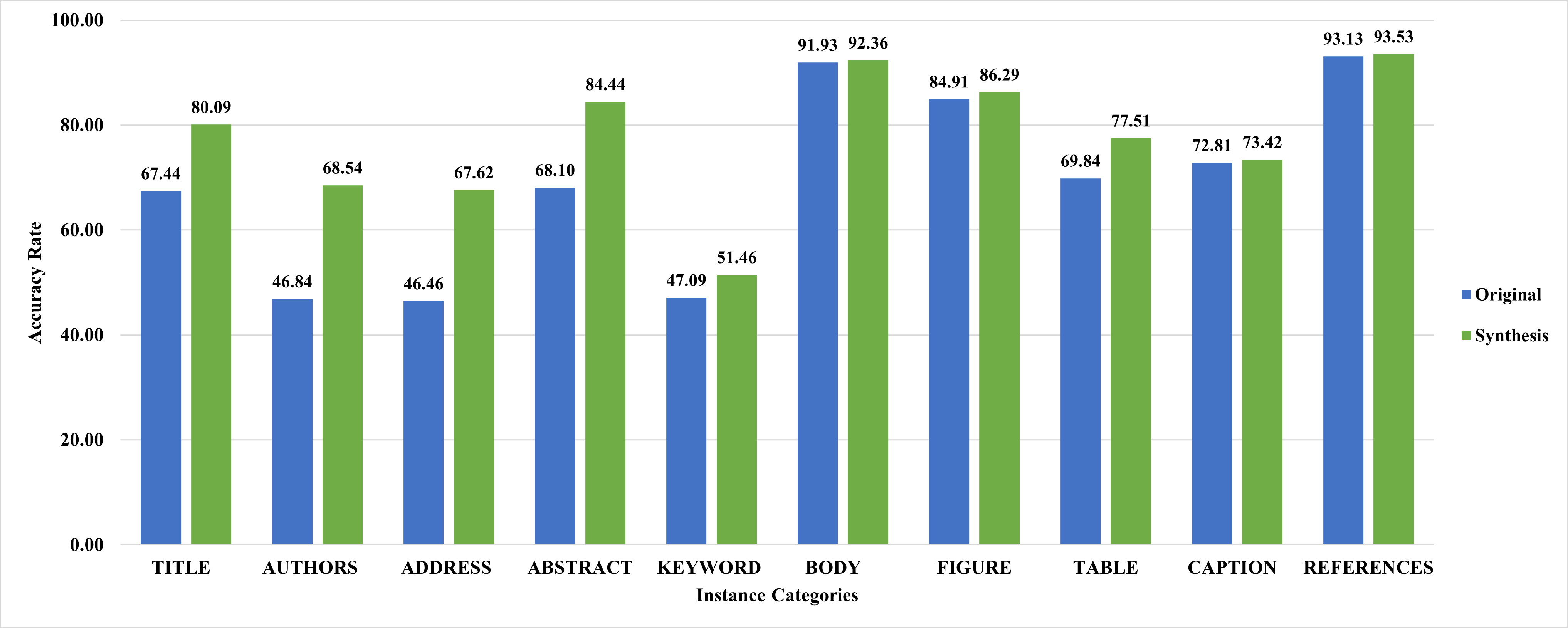}
    \end{center}
\caption{Detection results comparison between two data sets by labels at 0.5 IoU}
\end{figure}

Our framework performs well on major region detection from scientific literature (Fig. 6). However, it still fails to detect non-rectangle region in the images of scientific documents (left figure in Figure 8). We attribute this to the restriction of annotation methods. For data set annotation, we used rectangles instead of bounding polygons, which could be a viable and more flexible representation for non-rectangular regions. Moreover, duplicated bounding boxes are found to detect a single region area in the images of scientific documents (right figure in Fig. 8). We think this issue will be improved by training by deeper network, such as VoVNetV2-99., and with more epochs.

\begin{figure}[h!]
    \begin{center}
        \includegraphics[width=0.5\textwidth]{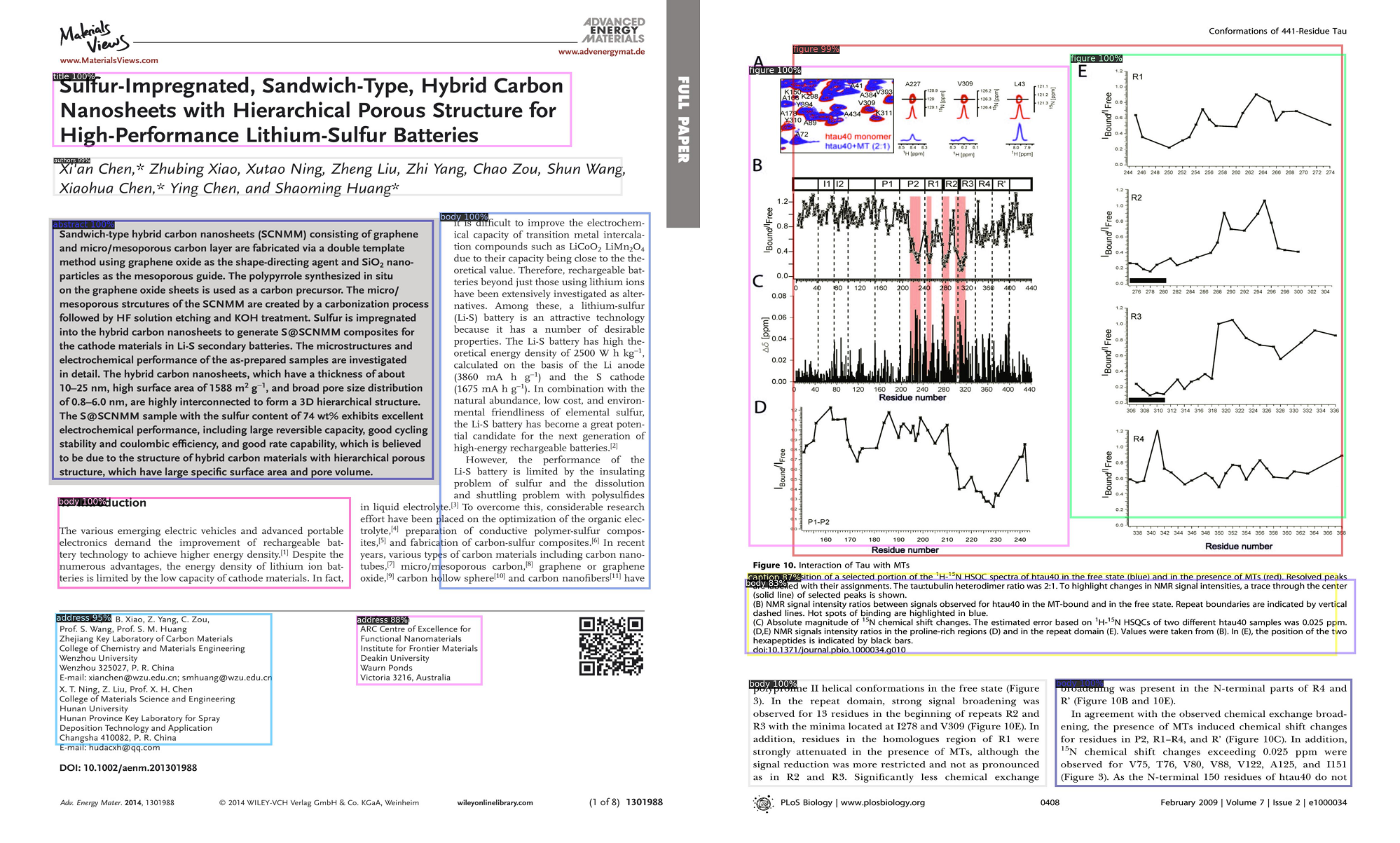}
    \end{center}
\caption{Examples of failure: left figure represents failure of locating polygon regions, and right figure represents overlap of bounding boxes for single region}
\end{figure}

\section{Conclusion}
For SLLD tasks, we introduce a novel end-to-end learning and vision-based framework. The major regions in scientific literature documents will be detected through the model which is trained by our framework. This model not only detects text regions but also figures and tables. Our approach is easy to adapt and implement for a broad range of scientific literature formats and domains, since it does not require extraction of additional features (e.g. text information of document). Fine-tuning pre-trained model which is generated from irrelevant tasks is feasible through our experiment. Specifically, we used a pre-trained model with MS COCO \cite{b22} which does not have any classes in the data set that are related to our SLLD work. Compared with other approaches of document layout analysis that are only working on some parts of the documents, our approach is built on the entire document rather than some regional information, and therefore it offers a possible way to construct large corpus for downstream NLP. 

Regarding the continuing work, more instances of minority classes should be added into training data set, such as title, authors, keyword, to improve minority object detection results. Although we apply transfer learning to our models, the training process is still not as efficient as what we expected. We are going to replace complex network with lite neural network, such as MobileNetV2.

\end{document}